%% file: acl2021.tex
\documentclass[11pt,a4paper]{article}
\usepackage{authblk}
\usepackage[hyperref]{acl2021}
\usepackage{times}
\usepackage{latexsym}
\usepackage{graphicx}
\usepackage{multirow}
\usepackage{threeparttable}
\usepackage{float}

\usepackage{booktabs}
\usepackage[normalem]{ulem}
\useunder{\uline}{\ul}{}

\usepackage{array}
\newcolumntype{P}[1]{>{\centering\arraybackslash}p{#1}}

\hypersetup{colorlinks=true,breaklinks=true}

\aclfinalcopy 

\title{ViT5: Pretrained Text-to-Text Transformer for Vietnamese Language Generation}

\author{Long Phan$^{1,2}$, Hieu Tran$^{1}$, Hieu Nguyen$^{1,2}$, Trieu H. Trinh$^{1,3}$ \\
        $^{1}$VietAI Research \\  $^{2}$Case Western Reserve University \\
        $^{3}$New York University \\
        \texttt{long.phan@case.edu}}

\date{}
\begin{document}
\maketitle
\begin{abstract}

\input{sections/abstract}
\end{abstract}

\section{Introduction}
\input{sections/introduction}

\section{Related Work}
\input{sections/related_works}


\section{ViT5}

\input{sections/vit5}
\input{table/data}


\section{Abstractive Summarization}
\input{sections/datasets}
\input{table/wikilingua}

\input{sections/experiments}

\input{sections/results}


\section{Named Entity Recognition (NER)}
\input{sections/ner}

\section{Discussion}
\input{sections/discussion}
\section{Conclusion}
\input{sections/conclusion}

\section{Acknowledgements}
We would like to thank the Google TPU Research Cloud (TRC) program and VietAI for providing us with free access to TPU v3-8 to train and finetune large ViT5 models. 
\newpage

\bibliographystyle{acl_natbib}
\bibliography{anthology,acl2021}


\end{document}

%% file: sections/abstract.tex
We present ViT5, a pretrained Transformer-based encoder-decoder model for the Vietnamese language. 
With T5-style self-supervised pretraining, ViT5 is trained on a large corpus of high-quality and diverse Vietnamese texts. We benchmark ViT5 on two downstream text generation tasks, Abstractive Text Summarization and Named Entity Recognition. Although Abstractive Text Summarization has been widely studied for the English language thanks to its rich and large source of data, there has been minimal research into the same task in Vietnamese, a much lower resource language. In this work, we perform exhaustive experiments on both Vietnamese Abstractive Summarization and Named Entity Recognition, validating the performance of ViT5 against many other pretrained Transformer-based encoder-decoder models. Our experiments show that ViT5 significantly outperforms existing models and achieves state-of-the-art results on Vietnamese Text Summarization. On the task of Named Entity Recognition, ViT5 is competitive against previous best results from pretrained encoder-based Transformer models. Further analysis shows the importance of context length during the self-supervised pretraining on downstream performance across different settings.

%% file: sections/introduction.tex
In recent years, Transformer-based architecture models and pretrained language models (LMs) have played a crucial role in the development of Natural Language Processing (NLP). Large pretrained models such as ELMo \cite{elmo}, GPT \cite{GPT}, BERT \cite{bert} is trained on large corpora and have the ability to derive contextual representation of the language(s) in the training data. After pretraining is complete, these models achieved state-of-the-art results on a broad range of downstream tasks \cite{bert}. These self-supervised learning methods make use of learning objectives such as Masked Language Modeling (MLM) \cite{bert} where random tokens in the input sequence are masked and the model attempts to predict the original tokens. The successes of pretrained models in English have inspired new research efforts to develop pretrained models in other languages such as Vietnamese (i.e., PhoBERT \cite{phobert} and ViBERT \cite{vibert}) and Italian \cite{it5}. There are also ongoing efforts to develop multilingual pretrained models (mT5 \cite{mT5}, mBART \cite{mbart}), in order to improve performance across multiple languages by learning both general and language-specific representations.

A short time ago, BARTpho \cite{bartpho}, a large pretrained sequence-to-sequence model for Vietnamese inheriting BART style \cite{bart}, demonstrated the effectiveness of pretrained language models on Vietnamese abstractive summarization. Nevertheless, there are some past works that have shown that T5 architecture \cite{t5} might outperform BART in some aspects (i.e., \cite{phan-etal-2021-cotext}).
Inspired by that, we propose ViT5, trained on the Vietnamese monolingual subset of CC100, following the architecture and training methodology in \citet{t5}. We perform exhaustive comparisons on downstream performance to many different pretrained Transformer-based models \cite{nguyen2021viesum, bartpho, to2021monolingual}. Specifically, we finetune the ViT5 on two summarization datasets, Wikilingua \cite{wikilingual} and Vietnews \cite{vietnews}, and one Named Entity Recognition dataset (PhoNER \cite{PhoNER_COVID19}).

Text summarization is an important downstream task whose input is a free-form text paragraph or document(s), and the output sequence is expected to be a short summarization of the input. ViT5 achieves state-of-the-art results on both two of the single-document summarization tasks. We also perform an analysis on the max-length hyper-parameter for input and output sequences during self-supervised learning and showed that longer lengths that match the downstream document's length lead to better result.

For NER, we reformulated the per-token classification task into a generation task, where the decoder reconstructs the original input sentence with inserted Named Entity tags following each token \cite{scifive}. This simple and straightforward formulation achieves competitive results in comparison to direct per-token classification done on encoder-only model~\cite{phobert}.
\label{introduction}

%% file: sections/related_works.tex
There are lots of abstractive summarization studies in English. In an early example, \cite{gehrmann-etal-2018-bottom} employed a bottom-up content selector (BottomUp) to determine which phrases in the source document should be part of the summary, and then a copy mechanism was applied only to pre-select phrases during decoding. Their experiments obtained significant improvements on ROUGE for some canonical summarization datasets. 

In recent years, pretrained language models have been used to enhance performance on language generation tasks. \cite{Liu2019TextSW} developed a Transformer-based encoder-decoder model so that pretrained language models like BERT can be adopted for abstractive summarization. Here, the authors proposed a novel document-level BERT-based encoder (\textit{BERTSum}) and a general framework encompassing both extractive and abstractive summarization tasks. Based on \textit{BERTSum}, \citet{dou2021gsum} introduced \textit{GSum} that effectively used different types of guidance signals as input in order to generate more suitable words and more accurate summaries. This model accomplished state-of-the-art performance on four popular English summarization datasets.

Meanwhile, there are a small number of studies on Vietnamese text summarization. Most of these focus on inspecting extractive summarization. The researchers \cite{8573420} compared a wide range of extractive methods, including unsupervised ranking methods (e.g., LexRank, LSA, KL-divergence), supervised learning methods using TF-IDF and classifiers (e.g., Support Vector Machine, AdaBoost, Learning-2-rank), and deep learning methods (e.g., Convolutional Neural Network, Long-Short Term Memory). Similarly, the authors \cite{vietnews} also evaluated the extractive methods on their own dataset, which was released publicly as a benchmark for future studies.

Recent work \cite{quoc2021monolingual} investigated the combination of a pretrained BERT model and an unsupervised K-means clustering algorithm on extractive text summarization. The authors utilized multilingual and monolingual BERT models to encode sentence-level contextual information and then ranked this information using the K-means algorithm. Their report showed that monolingual models achieved better results compared when to multilingual models performing the same extractive summarization tasks. However, due to the lack of studies on Vietnamese abstractive summarization, we compare both multilingual and monolingual encoder-decoder models. 

%% file: sections/vit5.tex
In this section, we will explain our newly released ViT5 models, the vocabulary generation steps, the pretraining data, and the training setup. 

\begin{figure}[H]
    \centering
    \includegraphics[width=0.5\textwidth,height=0.5\textheight,keepaspectratio]{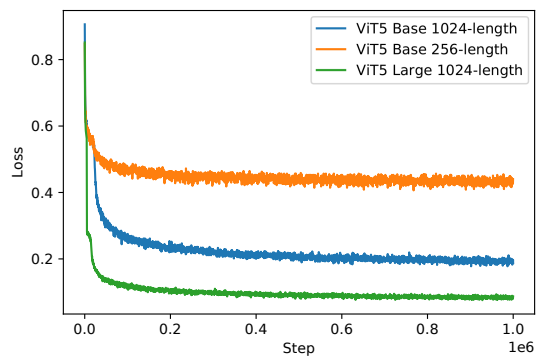}
    \caption{Loss curves for the masked span prediction task were used to pretrain the ViT5 models. Larger model with larger context optimizes much better, which leads to better downstream performance.}
    \label{fig:loss}
\end{figure}

\begin{figure*}[hbt!]
    \centering
    \includegraphics[width=0.95\textwidth,height=0.55\textheight,keepaspectratio]{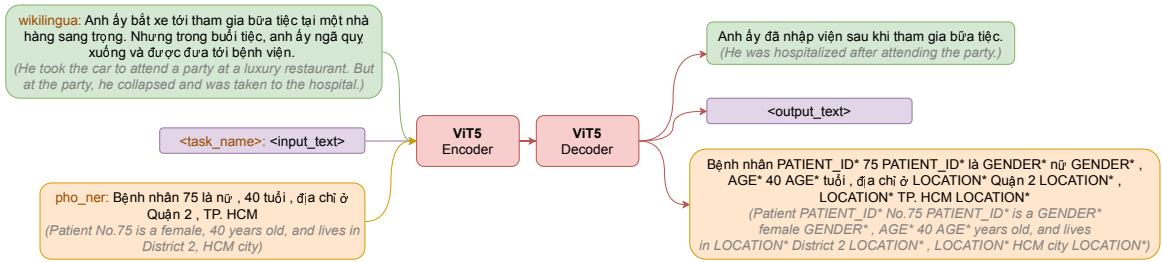}
    \caption{An overview of ViT5 encoder-decoder architecture, with input-output examples of two downstream tasks. For Named Entity Recognition, the decoder reconstructs the sentence with inserted Entity tags.}
    \label{fig:vit5}
\end{figure*}

\subsection{Model}
ViT5 follows the encoder-decoder architecture proposed by \citet{attention} and the T5 framework proposed by \cite{t5}. The original works of T5 proposed five different configs of model size: small, base, large, 3B, and 11B. For the purpose of practical study, we adapt the base (310M parameters) and large (866M parameters) models for ViT5 models and leave bigger models for future works. 

We train ViT5 models with two different input and output lengths: 256 and 1024-length. We thoroughly experimented with these two models to have an insight into the importance of pretraining data length for summarization tasks. For the self-supervised training learning objectives, we use the span-corruption objective with a corruption rate of 15\%. Figure \ref{fig:loss} shows the computed loss during the self-supervised training stage for the three models.

\subsection{Vocabulary}
Different from some other current Vietnamese Transformer-based language models, we find that an effective vocabulary can contribute a significant improvement to our model performance. Therefore, we did pre-process on a 5GB subset of our pretraining corpus with care like normalizing punctuation and capitalization, splitting numbers. We fixed the size of vocabulary to 36K sub-words and trained SentencePiece \cite{Kudo2018SentencePieceAS} model on that dataset.

\subsection{Pretraining Data}
We use the CC100 Dataset (Monolingual Datasets from Web Crawl Data) \cite{ccnet, conneau-etal-2020-unsupervised}. The corpus contains monolingual data for over 100 languages. The corpus was constructed using the pipeline provided by \cite{ccnet} through processing January-December 2018 Commoncrawl snapshots. The total size for the Vietnamese Corpus is 138GB of raw text. We process and filter out 69GB of short paragraphs for 256-length model and 71GB of long paragraphs for 1024-length model.


%% file: table/data.tex
\begin{table}[hbt!]
\centering
\caption{Input and Output Length of Finetuned Datasets}
\begin{tabular}{|c|c|l|l|c|l|l}
\hline
\multicolumn{1}{|l|}{}               & \multicolumn{3}{c|}{Wikilingua}          & \multicolumn{3}{c|}{Vietnews}                                    \\ \hline
Train     & \multicolumn{3}{c|}{13707}                 & \multicolumn{3}{c|}{99134}                                         \\ \hline
Test     & \multicolumn{3}{c|}{3916}                 & \multicolumn{3}{c|}{22498}                                         \\ \hline
\#avg body length     & \multicolumn{3}{c|}{521}                 & \multicolumn{3}{c|}{519}                                         \\ \hline
\#avg abstract length & \multicolumn{3}{c|}{44}                  & \multicolumn{3}{c|}{38}                                          \\ \hline
\end{tabular}
\label{data}
\end{table}

%% file: sections/datasets.tex
\subsection{Wikilingua}
Wikilingua \cite{wikilingual} is a large-scale multilingual corpus for abstractive summarization tasks. The corpus consists of 18 languages, including Vietnamese. These article and summary pairs are extracted from WikiHow\footnote{https://www.wikihow.com}. These articles have been reviewed by human authors to ensure quality. The Vietnamese articles are translated from the original English articles and have been reviewed by WikiHow's international translation team.

\subsection{Vietnews}
Vietnews \cite{vietnews} is a single-document abstractive summarization dataset including news data from reputable Vietnamese news website (\textit{tuoitre.vn}, \textit{vnexpress.net}, and \textit{nguoiduatin.vn}). 
The authors of this work removed all articles related to questionnaires, analytical comments, and weather forecasts to ensure the quality of document summarization. The final released dataset only includes long document news events. The data consists of 150704 word-level news articles with a summary abstract and body text pairs. We follow the filtering pipeline by \citet{bartpho} to deduplicate the train/dev/test dataset. The statistics after filtering are shown in Table \ref{data}.

%% file: table/wikilingua.tex
\begin{table*}[ht]
\centering
\caption{Test result on Wikilingua and Vietnews Summarization}

\begin{threeparttable}
\begin{tabular}{l|lll|lll}
\hline
\multirow{2}{*}{Models}                                         & \multicolumn{3}{c|}{WikiLingua}                  & \multicolumn{3}{c}{Vietnews}                     \\ \cline{2-7} 
                                                                & ROUGE-1        & ROUGE-2        & ROUGE-L        & ROUGE-1        & ROUGE-2        & ROUGE-L        \\ \hline
\begin{tabular}[c]{@{}l@{}}Transformer \\ (RND2RND)\end{tabular} & 46.25          & 16.57          & 29.82          & 57.56          & 24.25          & 35.53          \\ \hline
PhoBERT2PhoBERT                                                 & 50.4           & 19.88          & 32.49          & 60.37          & 29.12          & 39.44          \\ \hline
mBERT2mBERT                                                           & 52.82          & 20.57          & 31.55          & 59.67          & 27.36          & 36.73          \\ \hline
mBART                                                           & 55.21          & 25.69          & 37.33          & 59.81          & 28.28          & 38.71          \\ \hline
mT5                                                             & 55.27          & 27.63          & 38.30           & 58.05          & 26.76          & 37.38          \\ \hline
BARTpho                                                       & \textcolor{gray}{57.16}          & \textcolor{gray}{31.18}          & \textcolor{gray}{40.89}          & 61.14          & 30.31          & 40.15          \\ \hline
\hline
ViT5\textsubscript{base 256-length}                                            & 57.86          & 29.98          & 40.23          & 61.85          & 31.70          & 41.70          \\ \hline
ViT5\textsubscript{base 1024-length}                                           & {\ul 58.61}    & {\ul 31.46}    & {\ul 41.45}    & {\ul 62.77}    & {\ul 33.16}    & {\ul 42.75}    \\ \hline
ViT5\textsubscript{large 1024-length}                                          & \textbf{60.22} & \textbf{33.12} & \textbf{43.08} & \textbf{63.37} & \textbf{34.24} & \textbf{43.55} \\ \hline
\end{tabular}
\begin{tablenotes}
      \small
      \item \textit{Notes:} The best scores are in bold and second best scores are underlined. The scores in gray color are our experiments. Code and models for reproducing our experiments: https://github.com/vietai/ViT5

\end{tablenotes}
\end{threeparttable}
\label{wikilingua}

\end{table*}

%% file: sections/experiments.tex
\subsection{Baselines}

In order to verify the effectiveness of our proposed methods, we compare ViT5 models with the Transformer models based on \cite{attention}, the ViSum BERT2BERT models \cite{nguyen2021viesum}, multilingual encoder-decoder model \cite{mT5, mbart}, and Vietnamese encoder-decoder BARTpho model \cite{bartpho}. The baseline transformer models (labeled RND) have a multi-head self-attention and a feed-forward network. RND models are initialized with random weights. 
For the BARTpho models, we follow the models set up and results released by \cite{bartpho}. All finetuned ViT5 models are conducted with a sequence length of 1024.

%% file: sections/results.tex
\subsection{Results}
\label{wikilingua_task}
We report the results of the ViT5 models on two datasets: Wikilingua and Vietnews. We do experiments with two versions of pretraining ViT5: 256-length and 1024-length to have an insight into the importance of pretraining data's paragraph length for summarization in Vietnamese. We also compare the results of ViT5\textsubscript{base} and ViT5\textsubscript{large} models.

We use ROUGE (Recall-Oriented Understudy for Gisting Evaluation) as our benchmark metrics for both single document summarization datasets. The metric measures the overlap of n-grams and word sequences between two candidate and reference sequences. ROUGE-1, ROUGE-2, and ROUGE-L mean the overlap between unigram, bigram, and longest matching sequence, respectively.

\subsubsection{Wikilingua}

\label{vietnews_task}
The results of our models on Wikilingua summarization dataset are shown in Table \ref{wikilingua}. 
ViT5 models outperform all of the experimented pretrained models, achieving state-of-the-art on all ROUGE metrics. There is also a significant increase in ROUGE scores when the models are pretrained on a longer input and output sequence (1024 compared to 256). 

Both versions of ViT5\textsubscript{1024-length} achieve the highest results on Wikilingua summarization tasks across all ROUGE metrics with ViT5\textsubscript{large 1024-length} achieving state-of-the-art. There is a significant improvement in score between the base and large ViT5\textsubscript{1024-length} architectures (approximately 2\% for ROUGE-1, ROUGE-2, and ROUGE-L). This is predictable as the number of parameters of ViT5\textsubscript{large} (866M) is approximately 2.8 times larger than ViT5\textsubscript{base} (310M).

There are interesting results when comparing the results of 256-length and 1024-length versions of ViT5\textsubscript{base}. Although the finetuning settings are 1024-length for both ViT5\textsubscript{base} models, ViT5\textsubscript{base 1024-length} performs slightly better with ~1\% higher score for ROUGE-1, ROUGE-2, and ROUGE-L. These results are attributed to the longer sequences during self-supervised training. As reported in Table \ref{data}, the average words in an input body of Wikilingua corpus are more than 256 tokens, which can be considered long documents. For this reason, pretraining ViT5 on a 1024 sequence length corpus achieves better results on Wikilingua summarization task.

Two-out-of-three ViT5 models perform better than the published BARTpho model in summarizing Wikilingua corpus. This can be the result of the quality of pretraining data. While BARTpho (and PhoBERT) was trained on 20GB of news data, ViT5 models are trained using CC100, which is a subset of Common Crawl data. CC100 corpus contains more diverse and general representation of the Vietnamese language than news data. Meanwhile, Wikilingua is more of an academic or instruction representation than news-like text. 
 \subsubsection{Vietnews}
The size of Vietnews corpus is much larger than Wikilingua corpus (with 7.7\% for train and 5.8\% for test set). 
 The result of Vietnews abstractive summarization is in Table \ref{wikilingua}. Following the discussion of the need for an effective large pretrained encoder-decoder model in Section \ref{introduction}, we can see that there is a minimum increase in performance for the existing Vietnamese encoder-only model compared to the Transformer baseline. Pretraining on a large corpus of Vietnamese news, BARTpho still showed its limitation in the Vietnews summarization task with slightly better ROUGE scores than multilingual models (mBART and mT5).
 
 Our ViT5 models still achieve state-of-the-art on Vietnews task for both 256 and 1024-length. For a more specific news-domain corpus, ViT5 models achieve notable results on the news domain although being trained on a more general Vietnamese natural language domain (CC100). This supports the assumption that our ViT5 models learn a better representation of the Vietnamese language even for more domain-specific summarization problems. 
 
 Similar to the results discussed in Section \ref{wikilingua_task}, ViT5\textsubscript{base} models when pretrained on a longer sequence corpus (1024-length) achieve better performance in summarizing compared to a short sequence corpus (256-length) across all ROUGE metrics.
 The average input length for Vietnews documents is approximately the same as in the Wikilingua task (more than 500 words). Therefore, the quality of long sequences during self-supervised training data also leads to a better summarizing in downstream Vietnews finetuned tasks.


 \label{vietnews}

%% file: sections/ner.tex
\input{table/phoner_results}
To verify the effectiveness of ViT5 on classification tasks, we test our models on PhoNER\_COVID19 dataset \cite{PhoNER_COVID19}. PhoNER is a dataset for recognizing named entities related to the COVID19 domain in Vietnamese. The dataset consists of 35,000 entities in over 10,000 sentences. The goal is to recognize 10 entity types related to the domain of COVID19 and epidemics topics. The dataset was released and benchmarked with PhoBERT \cite{phobert}. 

We treat the NER classifications tasks as text-to-text generating tasks with tags of labels before and after an entity token \cite{scifive}. An example of NER in text-to-text format is shown in Figure \ref{fig:vit5}. The results are shown in Table \ref{phoner_result}.

The ViT5\textsubscript{large 1024-length} model, although effective in generating Vietnamese abstractive summarization, shows its limitation in classification tasks with lower F1 scores on NER task. On the other hand, our ViT5\textsubscript{base 1024-length} model still performs slightly better than PhoBERT\textsubscript{base} and competitively the same as the current state-of-the-art PhoBERT\textsubscript{large} on the PhoNER corpus.

%% file: table/phoner_results.tex
\begin{table}[H]
\centering
\caption{Test results on PhoNER\_COVID19}
\begin{threeparttable}

\begin{tabular}{p{4.5cm}|c}
\hline
Models             & Micro-F1      \\ \hline
XLM-R\textsubscript{large}       & 93.8          \\
PhoBERT\textsubscript{base}      & 94.2          \\
PhoBERT\textsubscript{large}      & \textbf{94.5}          \\
ViT5\textsubscript{base 256-length}  &      93.19         \\
ViT5\textsubscript{base 1024-length} & \textbf{94.5} \\
ViT5\textsubscript{large 1024-length} & 93.8 \\ \hline
\end{tabular}

\begin{tablenotes}
      \small
      \item \textit{Notes:} The best scores are in bold.
\end{tablenotes}
\end{threeparttable}
\label{phoner_result}

\end{table}


%% file: sections/discussion.tex
According to the results on both Wikilingua and Vietnews summarization tasks (Table \ref{wikilingua} and Table \ref{vietnews}), there is a steady increase in ROUGE scores going from the baseline Transformer, BERT2BERT related models (PhoBERT2PhoBERT and mBERT2mBERT), multilingual encoder-decoder models (mBART, mT5), to pretrained monolingual models (BARTpho and ViT5). For Vietnamese summarization tasks, monolingual encoder-decoder models noticeably outperform multilingual models, most likely thanks to their more focused and narrower pretraining stage.

Interestingly, a more general domain of pretraining texts can lead to a better domain-specific summarization performance. In Section \ref{vietnews_task}, our ViT5 models while being trained on a more general corpus (CC100), outperform current models that are trained on news-related corpus. More technical domains such as laws, medicals, or engineering are not tested as we leave these domain-specific summarization tasks for future studies.

The slightly better performance of ViT5\textsubscript{base 1024-length} compared to ViT5\textsubscript{base 256-length} suggests that longer document summarization (more than 512 tokens) need a comparatively longer context length during the pretraining stage.

%% file: sections/conclusion.tex

We introduce ViT5, a pretrained sequence-to-sequence Transformer model for the Vietnamese language. Leveraging the T5 self-supervised pretraining formulation on massive and high-quality Vietnamese corpora, we showed that finetuned ViT5 models are performant on both generation and classification tasks. We exhaustively compare ViT5 with other pretrained formulations on both multilingual and monolingual corpora. Our experiments show that ViT5 achieves state-of-the-art results on summarization in both Wikilingua and Vietnews corpus, and competitive results in generating Named Entity Recognition (NER) on the PhoNER\_COVID19 dataset. We also analyze and discuss the importance of context length during the self-supervised pretraining stage, which strongly influences and positively leads to better downstream performance.